\pdfoutput=1
\documentclass[11pt]{article}

\usepackage[margin=1in]{geometry}
\usepackage{times}
\usepackage{microtype}
\usepackage{graphicx}
\usepackage{subcaption}
\usepackage{booktabs}
\usepackage{multirow}
\usepackage{colortbl}
\usepackage{amsmath,amssymb}
\usepackage[utf8]{inputenc}
\usepackage{hyperref}
\usepackage[numbers]{natbib}
\usepackage{authblk}
\usepackage{xcolor}
\usepackage{colortbl}

\newcommand{\mae}[2]{#1 {\scriptsize $\pm$ #2}}
\newcommand{\bestmae}[2]{\textbf{#1 {\scriptsize $\pm$ #2}}}
\newcommand{\bestoverall}[1]{\cellcolor{gray!20}#1}

\providecommand{\Description}[1]{}

\title{Sub-City Real Estate Price Index Forecasting at Weekly Horizons Using Satellite Radar and News Sentiment}

\author[1]{Baris Arat}
\author[1]{Hasan Fehmi Ates}
\author[1]{Emre Sefer\thanks{Corresponding author.}}
\affil[1]{Ozyegin University, Istanbul, Turkey \\ \texttt{baris.arat@ozu.edu.tr, hasan.ates@ozyegin.edu.tr, emre.sefer@ozyegin.edu.tr}}
\date{}

\begin{document}
\maketitle

\begin{center}
\textit{Preprint. Under review.}
\end{center}

\begin{abstract}
Reliable real estate price indicators are typically published at city level and low frequency, limiting their use for neighborhood-scale monitoring and long-horizon planning. We study whether sub-city price indices can be forecasted at weekly frequency by combining physical development signals from satellite radar with market narratives from news text. Using over 350{,}000 transactions from Dubai Land Department (2015--2025), we construct weekly price indices for 19 sub-city regions and evaluate forecasts from 2 to 34 weeks ahead. Our framework fuses regional transaction history with Sentinel-1 SAR backscatter, news sentiment combining lexical tone and semantic embeddings, and macroeconomic context. Results are strongly horizon dependent: at horizons up to 10 weeks, price history alone matches multimodal configurations, but beyond 14 weeks sentiment and SAR become critical. At long horizons (26--34 weeks), the full multimodal model reduces mean absolute error from 4.48 to 2.93 (35\% reduction), with gains statistically significant across regions. Nonparametric learners consistently outperform deep architectures in this data regime. These findings establish benchmarks for weekly sub-city index forecasting and demonstrate that remote sensing and news sentiment materially improve predictability at strategically relevant horizons.
\end{abstract}

\noindent\textbf{Keywords:} time series forecasting, multimodal learning, urban analytics, remote sensing, sentiment analysis
\maketitle

\section{Introduction}\label{sec1}

Real estate markets are fundamental to economic stability and investment allocation, yet the price indices used to monitor them persistently suffer from temporal and spatial aggregation. Most widely used indices are published monthly or quarterly at national to city level~\cite{owusu2013,owusuansah2018propertyindex}. This restricts their utility for neighborhood-scale monitoring, timely risk assessment, and long-horizon urban planning. The forecasting literature has developed largely under these constraints. Much of the work relies on univariate time-series baselines or parametric multivariate econometric models that incorporate macroeconomic fundamentals, while more recent work increasingly explores machine learning approaches~\cite{plakandaras_forecasting_2015,larson_evaluating_2010,levantesi_importance_2020}.

Meanwhile, studies on property appraisal that integrate street imagery, satellite views, and textual descriptions outperform transaction-only baselines for individual property valuation~\cite{huang_multimodal_2025,tekouabou_ai-based_2024}. These findings suggest that physical and textual signals encode information not present in transaction records alone. Yet this multimodal paradigm has not transferred to index-level forecasting as much, where the prediction target is an aggregate time series rather than individual property values. This gap motivates our central question: can urban development patterns (observable from satellite imagery), combined with shifts in market sentiment (detectable in news text), improve long-horizon forecasting of sub-city price indices?

There is a strong conceptual basis for this hypothesis. Construction activity and infrastructure development alter the physical landscape before completed projects appear in transaction records, while news coverage of regulatory changes and market sentiment may signal shifts before they materialize in prices. Both sources therefore act as potential leading indicators. However, testing this hypothesis faces a data constraint. Satellite imagery with consistent global coverage from the Copernicus Sentinel missions extends back only to 2014. Combined with the low frequency of published indices, this leaves insufficient observations for robust multimodal evaluation.

\begin{figure}
\centering
\includegraphics[width=0.7\linewidth]{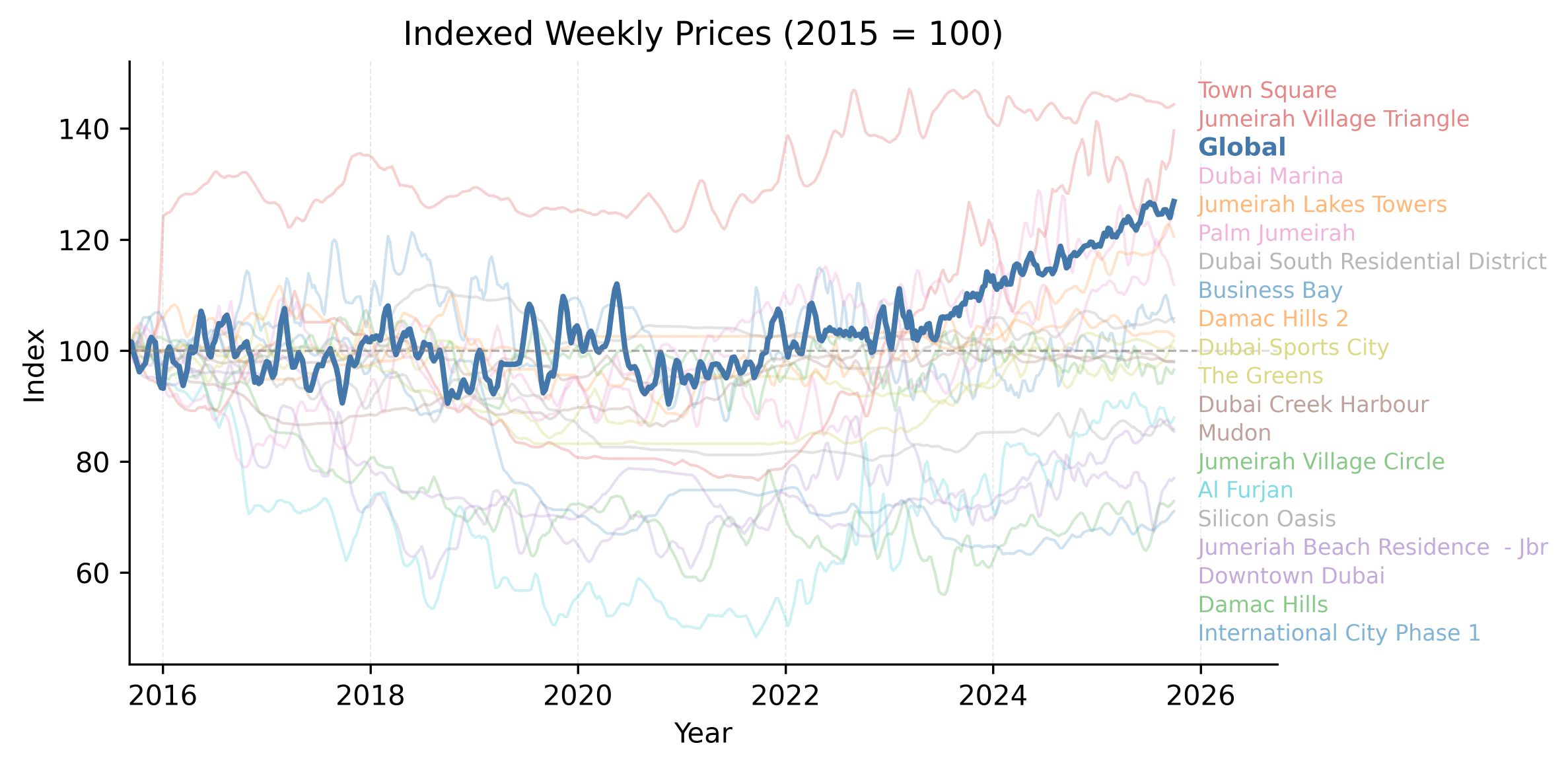}
\caption{Weekly price indices for 19 Dubai sub-city regions (2015--2025, rebased to 100). Regional trajectories diverge substantially from the citywide aggregate (bold).}
\Description{Time series plot showing 19 colored lines representing regional price indices diverging over time, with a bold dark blue line showing the citywide Global aggregate remaining more stable in the center.}
\label{fig:regional_divergence}
\end{figure}

We address this constraint by constructing weekly sub-city indices from granular transaction records. The Dubai Land Department (DLD) provides individual transaction data with date, location, and price since 2004~\cite{dld_opendata_real_estate_data}. This underlying transaction data retain sufficient granularity to support re-aggregation at finer temporal and spatial resolutions than published indices. We focus on 2015--2025 to align with Sentinel-1 availability and construct weekly price indices for 19 sub-city regions defined by master project boundaries. The resulting dataset contains approximately 500 weekly observations per region derived from over 350,000 transactions. As an exploratory analysis, Figure~\ref{fig:regional_divergence} illustrates clearly visible divergence from the citywide aggregate.

This experimental design offers both methodological and practical value. Methodologically, the combination of regions, weekly frequency, and decade-long coverage provides sufficient variation to evaluate multimodal contributions through systematic ablation while avoiding overfitting to a single aggregate series. Practically, sub-city weekly indices address a genuine information gap in regional dynamics which is relevant to developers, investors, and urban planners.

Our framework integrates four signal types: regional transaction history (prices and volumes), city-level news sentiment derived from GDELT, region-level Sentinel-1 synthetic aperture radar (SAR) backscatter, and macroeconomic context via interbank interest rates. For the textual modality, we combine lexical tone scores with semantic embeddings to capture topical structure beyond simple positive-negative polarity. For the spatial modaility, we prioritize Sentinel-1 SAR because, unlike optical imagery, radar supports consistent observation of built-up areas and highlights the structural and geometric changes associated with urban development~\cite{semenzato_mapping_2020}. Despite their widespread use in urban monitoring, such remote sensing signals have rarely been integrated into real estate price forecasting models. All data sources in this setting are publicly accessible so that this pipeline is applicable to other markets with similar transaction transparency, or partially applicable through globally available news sentiment and satellite imagery.

We evaluate forecasts across nine horizons from 2 to 34 weeks using rolling time-series cross-validation and benchmark multiple model families through systematic ablations. The results reveal a horizon-dependent modality contribution. At short horizons up to 10 weeks, transaction history alone performs comparably to multimodal configurations where autoregressive patterns dominate. Beyond 14 weeks, however, price-based models decay while multimodal configurations maintain accuracy. At long horizons, the full multimodal specification reduces mean absolute error by 35\% relative to price-only baselines, with improvements statistically significant across regions. Nonparametric learners such as K-nearest neighbors and random forest consistently outperform both linear models and recurrent neural networks in this data regime, likely because local similarity-based retrieval adapts naturally to regime shifts without requiring sufficient examples of each regime for parametric estimation.

These findings make four contributions:

\begin{itemize}
    \item We propose a multimodal framework for sub-city index forecasting that integrates transaction data with remote sensing, news sentiment, and macroeconomic indicators under a modular architecture that enables systematic ablation analysis.
    
    \item We establish empirical benchmarks for weekly long-horizon prediction using indices constructed from over 350,000 transactions across 19 Dubai regions, with evaluation across nine forecast horizons and multiple model families.

    \item We quantify modality contributions through ablation experiments. We identify the horizons at which sentiment and SAR improve forecasts and measure the magnitude of improvement.
    
    \item We provide methodological findings on feature design: semantic embeddings outperform lexical sentiment, SAR outperforms optical indices, and nonparametric learners outperform deep architectures in this regime.
\end{itemize}

\section{Related Work}\label{sec:related}

Real estate forecasting has evolved from univariate autoregressive methods toward models that incorporate exogenous predictors, though sub-city prediction remains underexplored. Early work established that aggregate house price indices can be forecast with meaningful accuracy over extended horizons using historical prices with macroeconomic predictors. Plakandaras et al.~\cite{plakandaras_forecasting_2015} apply ensemble empirical mode decomposition with support vector regression to the U.S. real house price index. Subsequent studies demonstrate that macroeconomic augmentation improves accuracy under specific conditions. Larson~\cite{larson_evaluating_2010} shows that error-correction models that incorporate income and rent outperform univariate specifications for California housing, particularly during the 2006 downturn when fundamental imbalances preceded price corrections. Levantesi and Piscopo~\cite{levantesi_importance_2020} find that population dynamics dominate in demand-driven markets such as London. Studies on the Dubai market have employed classical time-series methods at city level. ARIMA models applied to monthly indices~\cite{hepsen_forecasting_2011} and cointegration analyses that link prices to gold and macroeconomic variables~\cite{hepcsen2012relationship} establish baseline relationships but do not address sub-city dynamics. Recent work confirms that U.S. interest rate changes propagate to Dubai housing market activity~\cite{rashad_us_2024}. On the other hand, few studies attempt sub-city forecasting. Ge et al.~\cite{ge_integrated_2019} forecast monthly prices across subregion grids in New York and Beijing and find substantial cross-regional variation. Their results suggest that neighborhood-level dynamics contain information lost in city-level aggregation. This observation aligns with index construction literature that shows regional indices diverge systematically from citywide aggregates~\cite{bogin_missing_2016}.

Multimodal learning has been explored in both property appraisal and index-style forecasting, but the two settings differ in data availability and modality richness. Broader surveys of ML in real estate confirm that most studies still rely on structured tabular data from institutional sources, with few incorporating visual or textual modalities~\cite{tekouabou_ai-based_2024}. In property appraisal, recent work has explored incorporating descriptive text, interior imagery~\cite{hasan_multi-modal_2024} and neighborhood-level visual context using satellite imagery~\cite{kucklick_comparison_2021}, while an appraisal survey confirms added modalities improve accuracy across studies~\cite{huang_multimodal_2025}. On the index-style forecasting side, Wang~\cite{wang_housets_2025} addresses house price index forecasting at the ZIP-code level using aggregated regional price series. The study incorporates satellite imagery and applies large language models to generate textual descriptions of visual change, and reports mixed gains on predictive accuracy. Jiang et al.~\cite{jiang_modeling_2021} use a transformer encoder to represent neighborhood-level price history and combine it with Census features to predict multiclass hotspot tiers based on future price appreciation. Together, these studies show that additional signals beyond transaction records can be useful, although their value varies with the prediction target and aggregation level, consistent with the task-dependent nature of modality relevance in multimodal learning~\cite{liang_foundations_2024}.

Sentiment signals for real estate prediction can be derived from surveys or automated text analysis. Cheung et al.~\cite{cheung_effect_2021} construct investor and occupier sentiment indices from commercial property surveys and demonstrate predictive value for Australian property returns, with investor confidence particularly informative. Survey-based approaches offer reliability but are limited to low frequency and narrow geographic coverage. Recent work extracts sentiment directly from news. Xu~\cite{xu_real_2025} generates city-level sentiment indices using BERT-based classification of Chinese real estate news and shows that sentiment improves monthly vector autoregression forecasts, though effect sizes vary across cities and sources. Chen et al.~\cite{chen_research_2025} study high-frequency real estate sentiment from Chinese Weibo using a BERT-BiLSTM classifier and evaluate its predictive value with an ADL-MIDAS framework. These studies suggest that automated text sentiment can provide predictive information beyond price history, although the gains depend on the text source, the target variable, and the sampling frequency. However, current sentiment representations have limitations. Most approaches rely on lexicon-based polarity scores or domain-specific transformer sentiment models adapted to financial text~\cite{araci_finbert_2019}. Such methods may miss real estate specific semantic structure. Discussions of construction delays, regulatory changes, or infrastructure investments may carry predictive content that positive-negative classification cannot capture. Additionally, applications remain at monthly frequency.

Remote sensing provides direct observation of physical urban development. Sentinel-1 synthetic aperture radar offers consistent monitoring of built-up areas through strong backscatter responses that are robust to cloud cover and illumination constraints~\cite{torres_gmes_2012,semenzato_mapping_2020,verma_built-up_2023}. Time-series analyses show that Sentinel-1 captures meaningful urban structural change through built-up expansion and change detection~\cite{makineci_spatio}. Recent work also suggests that SAR contributes structural and geometric cues for built-up mapping that are not fully captured by optical spectral bands alone~\cite{li_extraction_2024}. These properties allow SAR to reflect construction activity and land transformation before completed structures appear in transaction records. Despite this potential, to our knowledge, SAR has seen limited adoption in real estate price forecasting.

An earlier conference version of this work explored short-horizon real estate forecasting using multi-context data and neural network models~\cite{arat2025multi}.

\section{Methods}\label{sec:methods}

\begin{figure*}
\centering
\includegraphics[width=\textwidth]{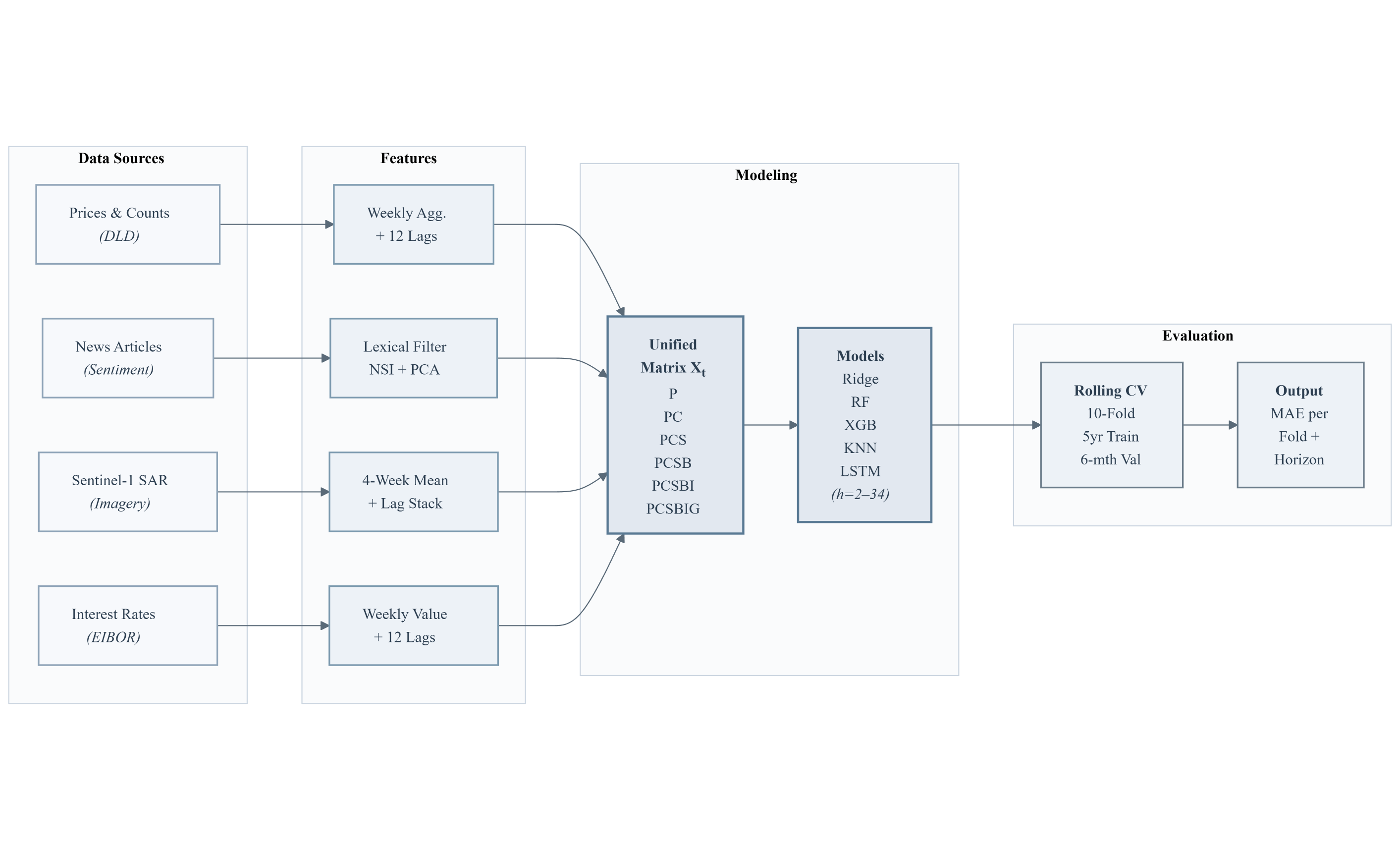}
\caption{Overview of the multimodal forecasting framework integrating transaction data, news sentiment, SAR imagery, and interest rates for sub-city price index prediction.}
\Description{Flow diagram showing data sources (DLD transactions, GDELT news, Sentinel-1 SAR, CBUAE interest rates) feeding into feature construction modules, which output to model training and cross-validation evaluation pipeline.}
\label{fig:methods_pipeline_overview}
\end{figure*}

This section describes the forecasting framework and the feature set. As a reference, Figure~\ref{fig:methods_pipeline_overview} summarizes the experiment pipeline and Table~\ref{tab:tags_config} defines modalities and tags. We construct a weekly panel of six feature sets: prices ($P$) and transaction counts ($C$) from DLD open data, news sentiment ($S$) from curated outlets based on GDELT, Sentinel-1 SAR backscatter ($B$) from Copernicus, interbank rates ($I$) from CBUAE, and citywide aggregates ($G$) derived from regional series. All series are aligned to Sunday-ending weekly calendar. Experiments take modular combinations of feature blocks through an additive tag system summarized in Table~\ref{tab:tags_config}. For any weekly series $x_{r,t}$, we define a causal lag block of length $L$ as:
\[
\mathbf{x}_{r,t}^{(L)}=\bigl(x_{r,t-1},\ldots,x_{r,t-L}\bigr)^\top.
\]
Transaction history enters via 12-lag blocks:
\[
\mathbf{P}_{r,t}=(P_{r,t-1},\ldots,P_{r,t-12})^\top \in \mathbb{R}^{12}, \quad
\mathbf{C}_{r,t}=(C_{r,t-1},\ldots,C_{r,t-12})^\top \in \mathbb{R}^{12},
\]
while interbank rates enter via $\mathbf{I}_{t}=(I_{t-1},\ldots,I_{t-12})^\top \in \mathbb{R}^{12}$. We append a citywide market context block:
\[
\mathbf{G}_{t}=\bigl(P^G_{t-1},\ldots,P^G_{t-12},\,C^G_{t-1},\ldots,C^G_{t-12}\bigr)^\top \in \mathbb{R}^{24}.
\]
News sentiment features $S_t\in\mathbb{R}^{15}$ and SAR features $\mathbf{B}_{r,t}\in\mathbb{R}^{15}$ are constructed in Sections~\ref{subsec:sentiment_data} and~\ref{subsec:sar_data}. All features are constructed from information available up to the end of week $t$. Since the minimum horizon is $h=2$, week-$t$ summaries do not leak future targets. Rolling statistics use trailing windows and winsorization thresholds are shifted by one period to exclude the current observation. At decision time $t$ and region $r$, the feature vector for a given modality tag $M \subseteq \{P,C,S,B,I,G\}$ is
\[
\mathbf{X}_{r,t}^{(M)} 
=
\bigl[\,
\mathbf{P}_{r,t}
\;\big|\;
\mathbf{C}_{r,t}
\;\big|\;
S_t
\;\big|\;
\mathbf{B}_{r,t}
\;\big|\;
\mathbf{I}_t
\;\big|\;
\mathbf{G}_t
\,\bigr],
\]
with blocks omitted when their modality is not in $M$. The features are horizon-invariant, while the prediction target varies with the forecast horizon: $y_{r,t}^{(h)} = P_{r,t+h}$. We evaluate nine forward horizons, $\mathcal{H} = \{2,\,6,\,10,\,14,\,18,\,22,\,26,\,30,\,34\}$ consisting of short, medium, and long-term forecasts.

\begin{table}
\centering
\footnotesize
\setlength{\tabcolsep}{6pt}
\renewcommand{\arraystretch}{1.15}
\caption{Additive modality tags and included feature blocks.}
\label{tab:tags_config}
\begin{tabular}{lp{0.65\linewidth}}
\toprule
\textbf{Tag} & \textbf{Included modalities (feature blocks)} \\
\midrule
P        & Prices ($12$ lags) \\
PC       & Prices ($12$) + Counts ($12$) \\
PCS      & Prices ($12$) + Counts ($12$) + Sentiment ($15$) \\
PCSB     & Prices ($12$) + Counts ($12$) + Sentiment ($15$) + SAR ($15$) \\
PCSBI    & Prices ($12$) + Counts ($12$) + Sentiment ($15$) + SAR ($15$) + Interest Rates ($12$) \\
PCSBIG   & Prices ($12$) + Counts ($12$) + Sentiment ($15$) + SAR ($15$) + Interest Rates ($12$) + Global context ($12{+}12$) \\
\bottomrule
\end{tabular}
\end{table}

\subsection{Datasets and Feature Constructions}

\subsubsection{Price Index and Volumes}

Regarding index construction methodology, median-based indices are simple and transparent, while repeat-sales and hedonic approaches aim to control for quality and compositional change but require stronger data support~\cite{hepcsen2012relationship}. We adopt the median approach because our sub-city regions lack the attribute consistency required for hedonic estimation at weekly frequency.

\begin{figure}
\centering
\includegraphics[width=0.4\linewidth]{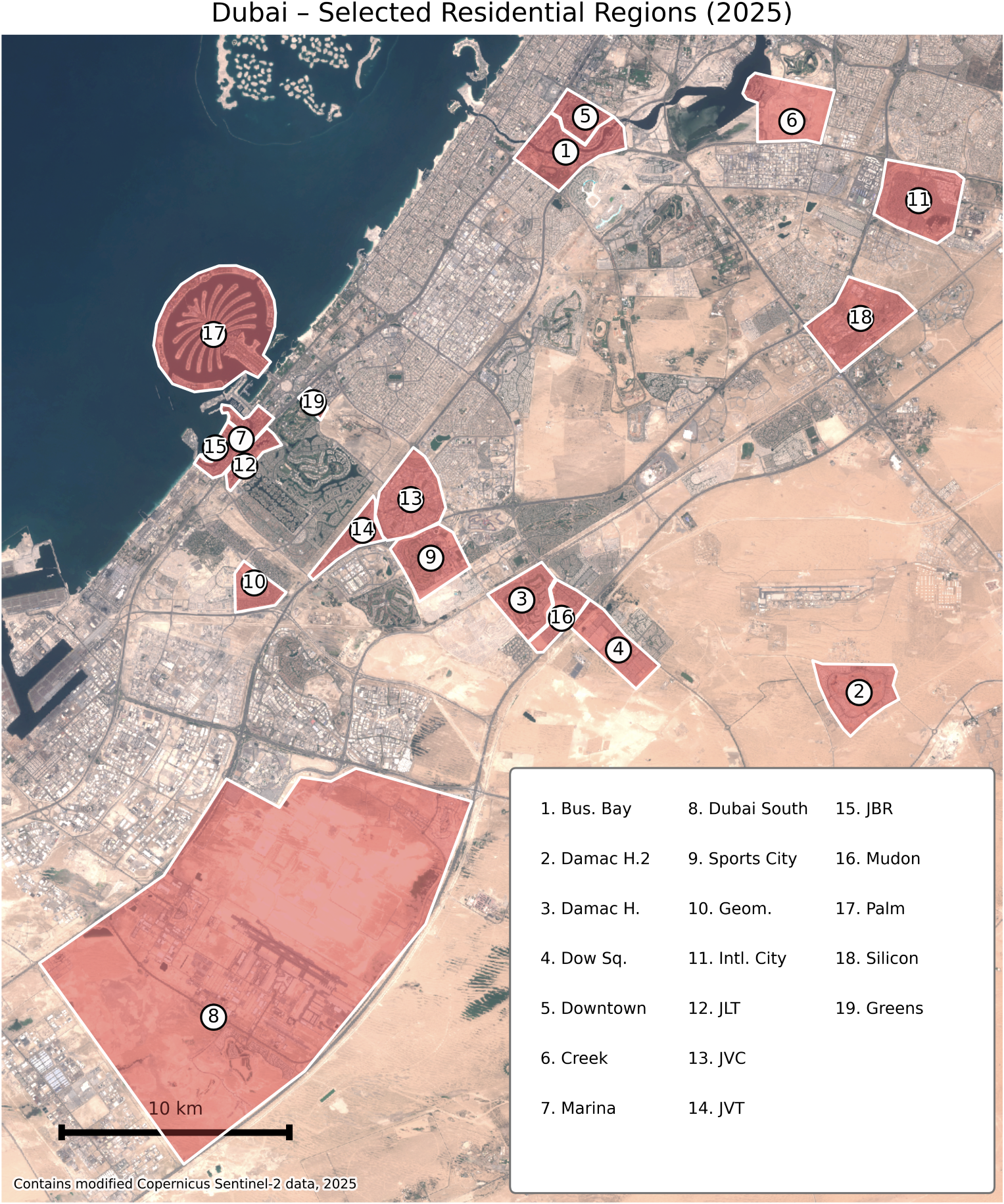}
\caption{Spatial distribution of selected Master Projects in Dubai.}
\Description{Map showing the geographic locations of Master Project regions across Dubai.}
\label{fig:dubai_regions_map}
\end{figure}

The weekly regional price index $P_{r,t}$ is derived from DLD transaction records containing transaction date, area, size, and Master Project based location labels. We focus on the 2015--2025 period and collect 19 Master Projects with consistent transaction history and spatial boundaries. We also construct a citywide ``Global'' aggregate by pooling all selected projects and use it only as a global context feature block rather than as an additional forecasting target. Project polygons were derived from public boundary definitions and manually checked for consistency with published district maps. Figure~\ref{fig:dubai_regions_map} shows the spatial distribution of selected regions. Filtering the raw data for the study window and selected projects yields over 350,000 transactions after standard outlier handling. For each region, index construction proceeds as follows. Per-$m^2$ transaction prices are first trimmed using fixed percentile cutoffs. A rolling median over the most recent 200 transactions is computed, the last value of each day is selected, and daily values are aggregated to weekly medians. The resulting series is aligned to a complete weekly grid and smoothed through forward filling of rare missing weeks, a trailing 4-week mean, winsorization of weekly changes exceeding $\pm3\times$ the 52-week rolling MAD. The final series is rebased to 100 at the first week. The Global series pools all selected projects and follows the same construction steps, using a larger rolling window of 600 transactions and winsorizing weekly changes at $\pm2\sigma$, where $\sigma$ is computed from a trailing 52-week window of weekly changes. The implications of alternative smoothing choices are examined separately through robustness diagnostics.

\begin{figure}[t]
\centering
\begin{subfigure}[t]{0.49\linewidth}
    \centering
    \includegraphics[width=\linewidth]{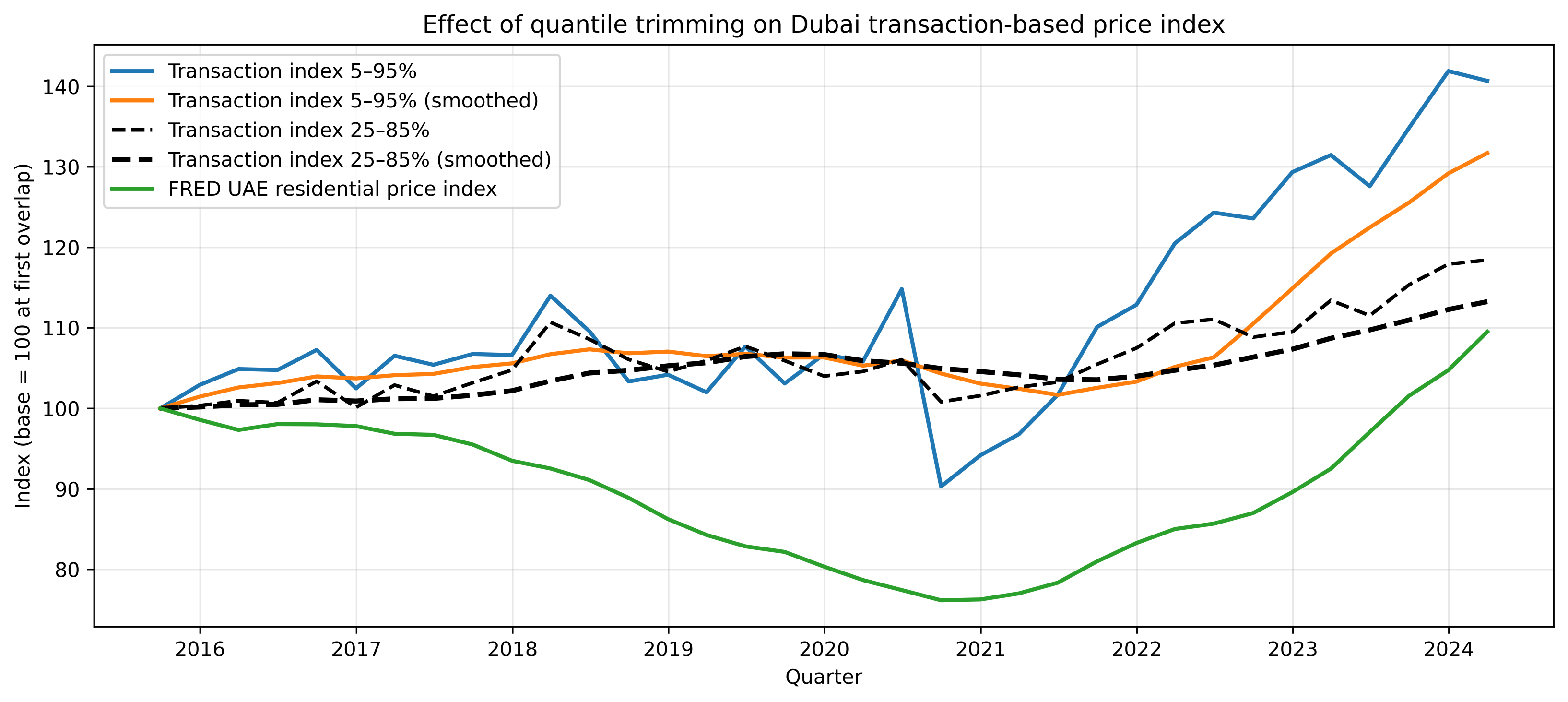}
    \caption{Citywide diagnostic: transaction-derived indices under alternative trimming compared to the macro benchmark.}
    \label{fig:quantile_trim_global}
\end{subfigure}
\hfill
\begin{subfigure}[t]{0.49\linewidth}
    \centering
    \includegraphics[width=\linewidth]{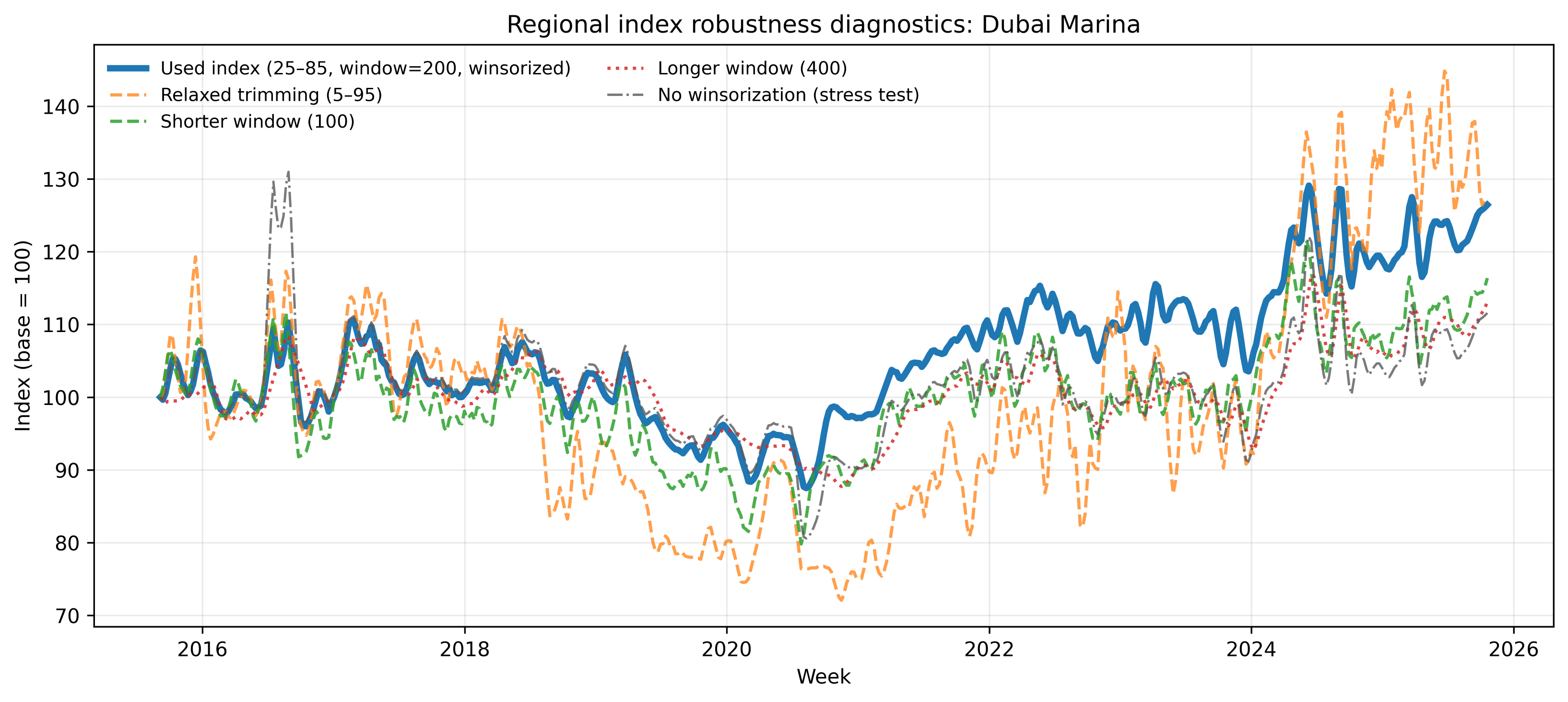}
    \caption{Regional diagnostic: a representative region illustrating stability differences under alternative indexing.}
    \label{fig:quantile_trim_region}
\end{subfigure}
\caption{Quantile trimming diagnostics for transaction-based index construction. Relaxed trimming ([5th, 95th]) improves citywide cycle fidelity relative to a macro benchmark, while more aggressive trimming ([25th, 85th]) yields a smoother regional monitoring signal with reduced sensitivity to idiosyncratic transactions.}
\Description{Two-panel figure. Left: citywide transaction-derived indices under two trimming schemes compared to the UAE property price index from FRED. Right: regional example showing the relaxed scheme is more volatile while the aggressive scheme is smoother.}
\label{fig:quantile_trim_diagnostics}
\end{figure}

\paragraph{Quantile trimming and robustness.}
Weekly sub-city indices constructed from transaction records are sensitive to heterogeneous unit mixes and occasional extreme transactions, particularly in regions with limited weekly volume. To obtain stable and comparable monitoring signals across regions, we apply percentile-based trimming to per-$m^2$ prices prior to aggregation, using fixed cutoffs estimated from the initial training window and applied uniformly thereafter. Figure~\ref{fig:quantile_trim_diagnostics} summarizes robustness diagnostics under two trimming schemes. At the citywide level (left), relaxed trimming ([5th, 95th]) aligns more closely with the UAE residential property price index from FRED (BIS)~\cite{fred_uae_bis}, while aggressive trimming ([25th, 85th]) yields a smoother long cycle. At the regional level (right), relaxed trimming induces higher volatility and deeper drawdowns, whereas aggressive trimming produces a smoother signal that follows a similar trajectory. Since our objective is cross-region stabilization and reliable weekly monitoring rather than appraisal-grade valuation, we adopt the [25th, 85th] trimming in the main analysis.

\subsubsection{News-Based Sentiment}
\label{subsec:sentiment_data}

We construct weekly city-level sentiment features from economic and real estate related news using a pipeline that combines source selection, domain filtering, and applies lexical tone and semantic embedding representation. We query GDELT GKG records~\cite{leetaru_gdelt_2013} from 2015 to 2025 for articles tagged with the housing theme. Source filtering restricts the corpus to English language outlets from a curated whitelist of 80 international and regional sources, including Gulf business press (e.g., \textit{Gulf News}, \textit{Khaleej Times}, \textit{Arabian Business}), major wire services (Reuters, AP), and established financial media (Bloomberg, CNBC, WSJ). We retrieve full article text using Trafilatura~\cite{barbaresi_trafilatura}, yielding approximately 27,000 articles with content. To ensure economic and real estate relevance, we apply keyword filtering using a comprehensive lexicon including macroeconomic terms (e.g., ``inflation,'' ``interest rate,'' ``monetary policy,'' ``GDP''), real estate terms (e.g., ``housing market,'' ``mortgage,'' ``construction,'' ``developer,'' ``vacancy''), and numerical markers. Articles lacking any lexicon match are excluded. For each retained article, we compute a Numerical Sentiment Index (NSI) from GDELT tone fields.
\[
  \mathrm{NSI}_i = 
  \frac{\mathrm{positive\_score}_i - \mathrm{negative\_score}_i}
       {\mathrm{positive\_score}_i + \mathrm{negative\_score}_i},
\]
Weekly NSI is computed as the median of article-level values within each week and smoothed using a 4-week rolling mean to reduce high-frequency noise. 

\begin{figure}[t]
\centering
\begin{subfigure}[t]{0.52\linewidth}
    \centering
    \includegraphics[width=\linewidth]{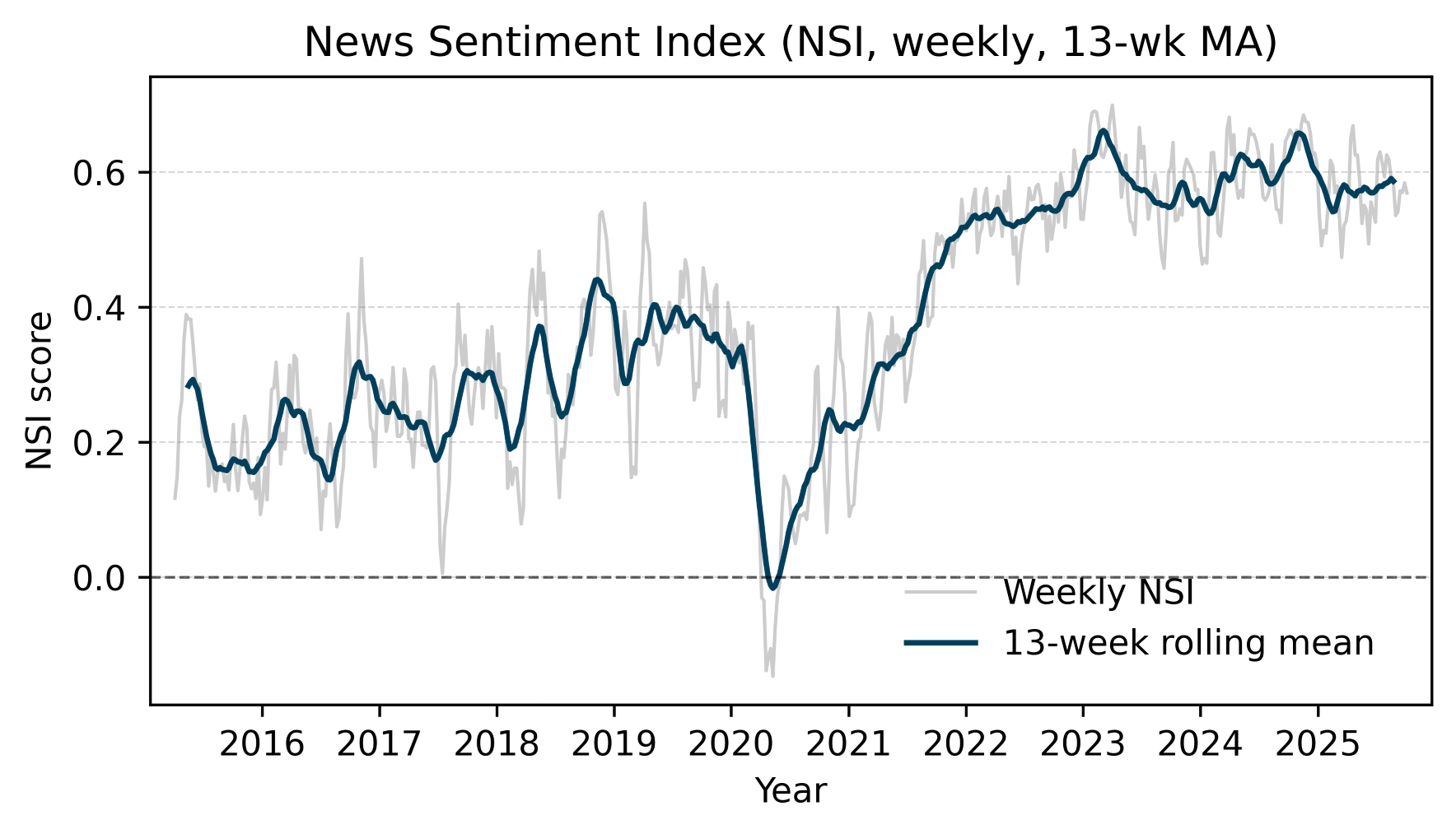}
    \caption{Weekly NSI, rolling mean. 13-week mean is for visualization only.}
    \label{fig:nsi_trend}
\end{subfigure}
\hfill
\begin{subfigure}[t]{0.44\linewidth}
    \centering
    \includegraphics[width=\linewidth]{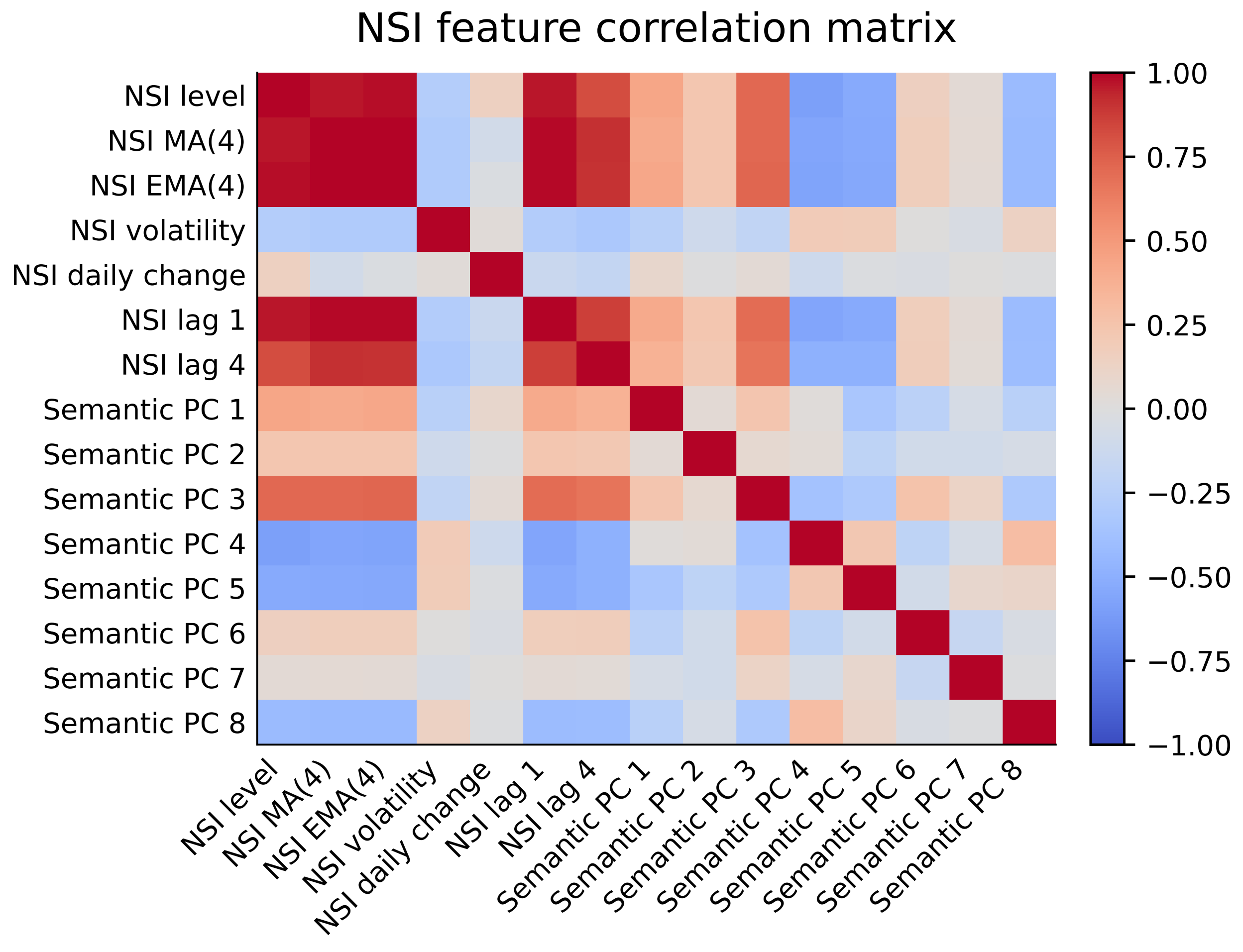}
    \caption{Lexical-semantic feature correlations.}
    \label{fig:nsi_feature_heatmap}
\end{subfigure}
\caption{Sentiment feature diagnostics. (a) Weekly NSI captures market sentiment shifts including the COVID-19 period. (b) Low correlation between lexical NSI features and semantic PCA components confirms complementary information.}
\Description{Left panel shows NSI time series with smoothed trend. Right panel shows correlation heatmap between lexical and semantic sentiment features.}
\label{fig:sentiment_diagnostics}
\end{figure}

We construct a 7-dimensional tone feature vector $s^{\text{tone}}_t = \bigl(\mathrm{NSI}_t,\;\Delta \mathrm{NSI}_t,\;\mathrm{vol}_t,\;\mathrm{MA}_t,\;\mathrm{EMA}_t,\;\mathrm{NSI}_{t-1},\;\mathrm{NSI}_{t-4}\bigr)\in\mathbb{R}^{7}$, where $\Delta\mathrm{NSI}_t$ is the weekly change, $\mathrm{vol}_t$ is the 4-week rolling standard deviation, and $\mathrm{MA}_t$ and $\mathrm{EMA}_t$ denote 4-week simple and exponential moving averages. To capture topical structure beyond lexical polarity, we encode article text using Sentence-BERT embeddings~\cite{reimers2019} with a MiniLM implementation~\cite{wang_minilm_2020}. Weekly embeddings are computed as the mean of article embeddings within each week, then reduced via PCA to 8 components. To avoid representation leakage, PCA is fit once using weekly embeddings from the initial training window of the rolling cross-validation procedure, and the learned projection is applied to all weeks without refitting. Component count is kept limited to prevent overfitting given the limited time series depth. Each component is exponentially smoothed ($\alpha=0.3$) to reduce week-to-week noise, and we denote the resulting vector by $\mathbf{pca}_t=(\mathrm{PC}_{t,1},\ldots,\mathrm{PC}_{t,8})\in\mathbb{R}^{8}$. The final weekly sentiment block concatenates lexical and semantic features as $S_t=\bigl[\,s^{\text{tone}}_t \mid \mathbf{pca}_t\,\bigr]\in\mathbb{R}^{15}$. Figure~\ref{fig:sentiment_diagnostics} shows the weekly NSI series alongside the correlation structure between lexical and semantic features. The NSI captures major market shifts, including sharp negative sentiment during early 2020 and following recovery. Lexical and semantic features exhibit low cross-correlation which suggests that components capture distinct narrative structure beyond aggregate tone.

\subsubsection{Satellite SAR Features}
\label{subsec:sar_data}

We use Sentinel-1 C-band SAR~\cite{torres_gmes_2012} because it captures meaningful urban structural change through consistent observations independent of cloud cover and illumination, which is critical for weekly monitoring~\cite{semenzato_mapping_2020, verma_built-up_2023, hafner_sentinel-1_2022,makineci_spatio}. Sentinel-1 GRD scenes (VV, VH) for 2015--2025 are retrieved via Google Earth Engine~\cite{gorelick_google_2017}, which provides calibrated, terrain-corrected backscatter in decibels. The resulting regional time series are resampled to a weekly grid and smoothed with a trailing 4-week mean to mitigate noise. Rare missing weeks in the base SAR series are addressed using forward filling only, and lagged SAR features are constructed strictly causally from past observations. To capture structural changes at multiple temporal scales, we augment the SAR statistics with lagged values at 4, 8, 12, and 20 weeks. These lags are chosen to span from short to medium-term development dynamics while ensuring full feature availability over the aligned study window. Let $\mathbf{b}_{r,t}^{(\ell)} \in \mathbb{R}^3$ denote the smoothed VV, VH, and VV/VH ratio at offset $t-\ell$. The SAR feature vector for region $r$ at week $t$ is $\mathbf{B}_{r,t} = \bigl(\mathbf{b}_{r,t}^{(0)},\,\mathbf{b}_{r,t}^{(4)},\,\mathbf{b}_{r,t}^{(8)},\,\mathbf{b}_{r,t}^{(12)},\,\mathbf{b}_{r,t}^{(20)}\bigr) \in \mathbb{R}^{15}$.

\subsubsection{Interest Rates}
\label{subsec:interest_rates}
EIBOR interbank rates (CBUAE)~\cite{cbuae_eibor_rates} are aggregated to weekly frequency and included via a 12-lag block $\mathbf{I}_t=(I_{t-1},\ldots,I_{t-12})$ to capture delayed monetary policy effects on housing markets~\cite{chen2022experimental,vonlanthen2023interest}.

\subsection{Forecasting Framework}
\label{subsec:forecasting_framework}

The forecasting task is to estimate the regional price index $y^{(h)}_{r,t}=P_{r,t+h}$ at horizon $h$ weeks ahead using multimodal explanatory inputs $\mathbf{X}_{r,t}^{(M)}$, where $M$ denotes a selected subset of modalities. Rolling time-series cross-validation is applied over 2015--2025 using 5-year training and 6-month validation windows advanced in 6-month steps. This procedure yields ten folds for all horizons except the longest $h{=}34$ weeks, for which nine folds are available because target observations end earlier. For each forecast horizon $h \in \mathcal{H}$, region $r$, and model $f$, we fit a separate predictor $f_{h,r}^{(M)}$ using features $\mathbf{X}_{r,t}^{(M)}$ and target $y^{(h)}_{r,t}=P_{r,t+h}$. Performance is evaluated using mean absolute error (MAE) on each validation window and reported as a macro-average across regions and folds:
\[
\text{MAE}_{\text{macro}}(h,f,M)
=
\frac{1}{R\,K_h}
\sum_{r=1}^{R}\sum_{k=1}^{K_h}
\text{MAE}_{r,k}(h,f,M),
\]
where $R=19$ and $K_h$ is the number of admissible CV folds for horizon $h$. We report $\text{MAE}(h)$ curves across horizons and fold-level standard deviations to assess robustness.

\subsubsection{Model Families}
\label{subsec:methods_models}

\textbf{Baselines.}
We include two univariate reference baselines operating on prices only.
The Naive-12Mean baseline forecasts region $r$ at horizon $h$ as the 12-week rolling mean of past prices:
\[
\hat{y}^{(h)}_{r,t,\text{Naive}} = \frac{1}{12}\sum_{i=1}^{12} P_{r,t-i}.
\]
As a classical parametric reference, we fit a non-seasonal ARIMA~\cite{box2015time} model on the price-only series (\texttt{P}). For each region and fold, the ARIMA order $(p,d,q)$ is selected on the fold’s training window by minimizing AIC over a bounded grid ($p \in \{0,\ldots,3\}$, $d \in \{0,1\}$, $q \in \{0,\ldots,3\}$). The ARIMA pipeline considers only converged fits with stationarity and invertibility enforced. Validation follows a rolling-origin protocol: for each decision week in the validation window, we refit ARIMA on all observations available up to that week using the selected order, then forecast forward and score the prediction at horizon $h$.

\textbf{Learning models.}
Beyond univariate baselines, we benchmark five learning model families of increasing flexibility. Ridge regression is fit with an $L_2$ penalty. Random Forest~\cite{breiman_random_2001} models use 300 trees with maximum depth 5 and minimum leaf size 1. XGBoost~\cite{chen2016xgboost} is implemented as a gradient-boosted tree ensemble trained on the same flat feature representation. A single fixed configuration is used across all regions, horizons, folds, and modality tags, with 600 trees, maximum depth 4, learning rate 0.05, subsample ratio 0.9, column subsample ratio 0.9, and $L_2$ regularization $\lambda=1.0$. K-Nearest Neighbors~\cite{cover_nearest_1967} regression uses $k=7$ with Manhattan distance on standardized feature. The LSTM~\cite{hochreiter_long_1997} operates on 12-step input windows aligned with the lag-based feature design and consists of three bidirectional LSTM layers with hidden size 32 and layer normalization, followed by two fully connected layers and a linear output head. Optimization is performed using AdamW~\cite{loshchilov_hutter_adamw_2019}.

\subsubsection{Statistical Testing}
\label{subsec:statistical_testing}

We assess the statistical significance of performance differences using nonparametric tests. All tests are conducted at $h \geq 26$ horizons, where multimodal effects are most pronounced. For each region and modality configuration, MAE is first averaged across the long-horizon set to obtain a single summary error per region. These region-level summaries constitute the paired samples used for statistical testing. First, within each model family, we apply the Friedman test to evaluate whether MAE differs across the six modality configurations (P, PC, PCS, PCSB, PCSBI, PCSBIG). If the Friedman test rejects the null hypothesis of equal performance ($p < 0.05$), we perform post hoc Wilcoxon signed-rank tests on marginal differences (e.g., PC$\rightarrow$PCS, PCS$\rightarrow$PCSB) to identify which modalities produce significant changes. Multiple testing is controlled using Bonferroni correction over the five tests performed for each modality addition. As a separate evaluation, we also test learning model differences. For this case, each model is represented by its best-performing modality configuration, and paired Wilcoxon signed-rank tests are conducted across regions to assess differences in best performances between models.

\subsubsection{Training and Implementation}
\label{subsec:implementation}

All experiments are implemented in Python using \texttt{scikit-learn} for Ridge, Random Forest, and KNN~\cite{pedregosa_scikit-learn_nodate}, \texttt{xgboost} for gradient-boosted trees~\cite{chen2016xgboost}, \texttt{statsmodels}\cite{seabold2010statsmodels} for ARIMA~\cite{box2015time} baselines, and \texttt{PyTorch} for the LSTM network~\cite{paszke_pytorch_2019}. Data preprocessing and visualization relied on \texttt{NumPy}~\cite{harris_array_2020}, \texttt{Pandas}~\cite{mckinney_data_2010}, and \texttt{Matplotlib}~\cite{hunter_matplotlib_2007}. Feature scaling is fit on training folds only to avoid leakage. Hyperparameters for Ridge, Random Forest, KNN, and the LSTM are selected via small grid searches on rolling CV splits using a subset of regions and horizons. ARIMA uses per-fold AIC-based order selection on training windows as described in Section~\ref{subsec:methods_models}, while XGBoost uses the fixed specification described here.

\section{Results and Discussion}
\label{sec:results}
\begin{figure} \centering \includegraphics[width=0.6\linewidth]{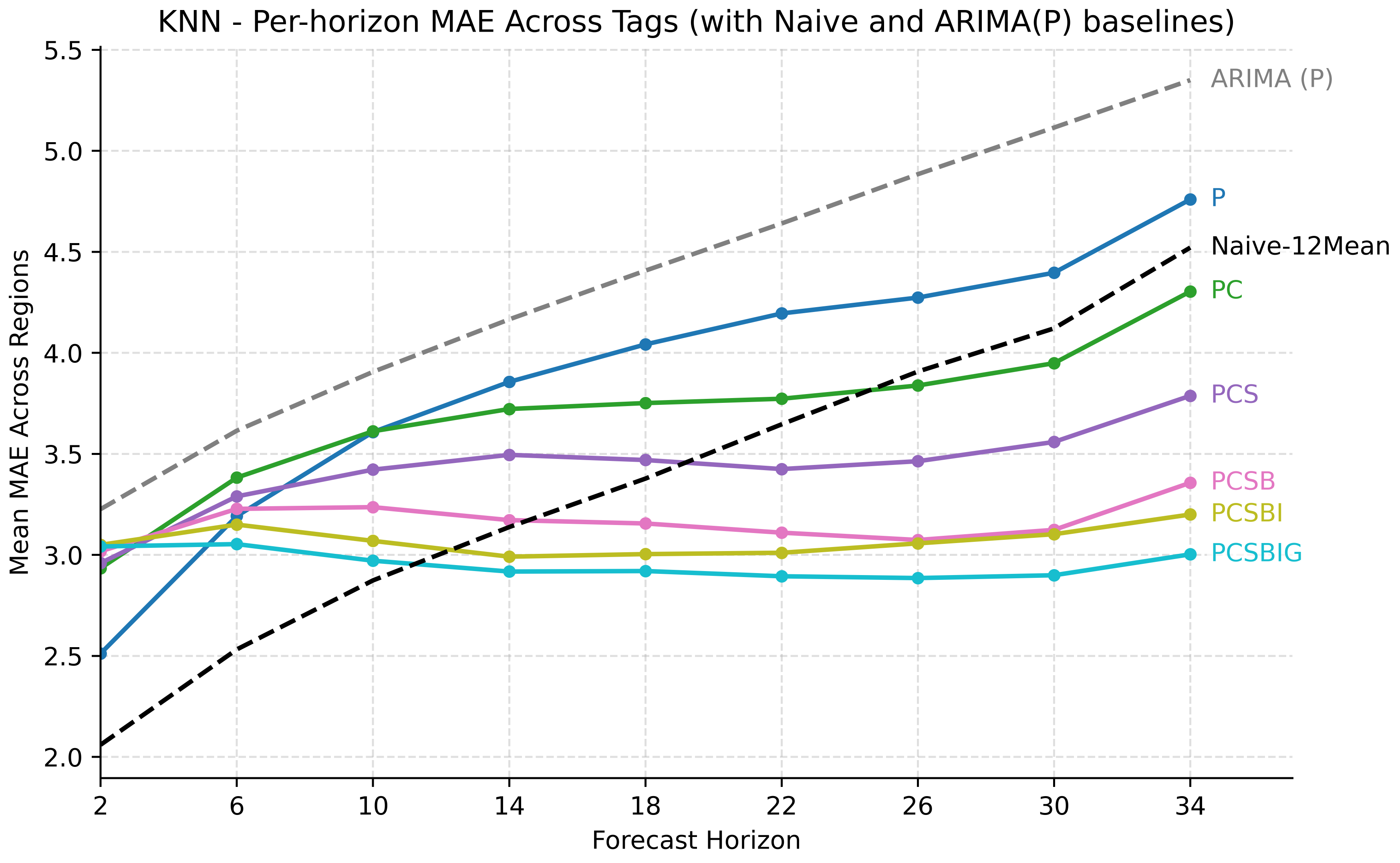}\caption{Per-horizon KNN performance by modality configuration. Lines show mean MAE across 19 regions. Beyond 14 weeks, multimodal configurations stabilize while price-only models deteriorate. Dashed lines show Naive-12Mean and ARIMA(\texttt{P}) baselines.} \Description{Line plot showing KNN MAE across forecast horizons for different modality configurations. All configurations cluster together at short horizons, but diverge beyond 14 weeks, with price-only showing highest error at long horizons.} \label{fig:knn_per_horizon} \end{figure}

The central finding is that sentiment and SAR provide limited benefit at short horizons but become the main driver for accurate long-horizon forecasting. Figure~\ref{fig:knn_per_horizon} illustrates this pattern on KNN. At horizons up to 10 weeks, the price-only KNN performs comparably to the best multimodal configurations, with MAE differences typically small relative to long-horizon gaps. Beyond 14 weeks, the price-only model loses predictive power steadily while multimodal configurations remain stable. Table~\ref{tab:grouped_all_models_by_tag_compact} summarizes MAE across models, modality configurations, and horizon groups. For KNN, long-horizon MAE decreases from 4.476 (prices only) to 2.929 (full multimodal), a 34.6\% reduction. Random Forest shows a similar pattern (4.310 to 3.260), and XGBoost is competitive (4.417 to 3.139). Ridge regression achieves the lowest short-horizon MAE but does not benefit from additional modalities. LSTM underperforms across all configurations, likely due to limited training data in the weekly regime. 

The univariate baselines provide useful reference points. Naive-12Mean performs competitively at short horizons but degrades rapidly as the forecast horizon increases, while ARIMA exhibits consistently higher error across all horizon groups. This behavior reveals the difficulty of long-horizon index forecasting using price history alone. 

\begin{figure*}
\centering
\includegraphics[width=\textwidth]{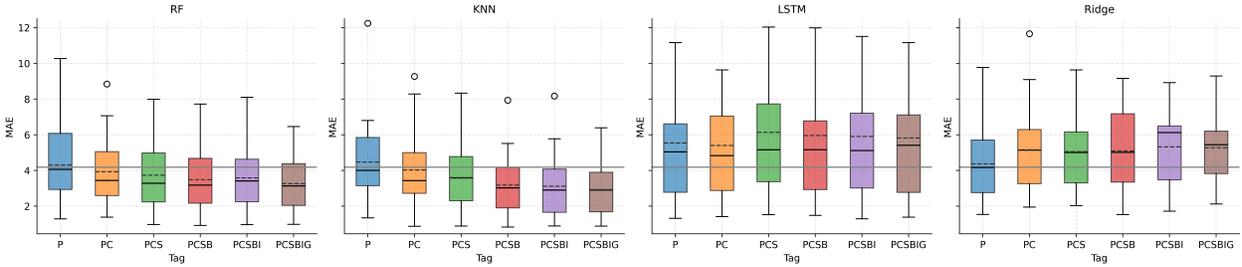}
\caption{Long-horizon MAE distributions across sub-city regions. Horizontal gray lines indicate Naive-12Mean baseline. KNN and Random Forest show the largest multimodal gains with tighter regional dispersion.}
\Description{Box plots comparing MAE distributions for four models across modality configurations. KNN and Random Forest show lower medians and less variance than Ridge and LSTM.}
\label{fig:main_boxplots}
\end{figure*}

Figure~\ref{fig:main_boxplots} shows long-horizon MAE distributions across regions. Nonparametric learners, KNN and Random Forest, achieve lower median errors and tighter dispersion, which indicates more consistent performance across regions under multimodal configurations. Direct comparison of these results with prior work is limited because existing studies using the same transaction dataset focus on different index constructions~\cite{hepsen_forecasting_2011}, methods such as macroeconomic cointegration~\cite{hepcsen2012relationship}, and unit-level prediction~\cite{elnaeem_balila_comparative_2024}. Our results therefore establish initial benchmarks for long-horizon, multimodal sub-city index forecasting.

\begin{table*}
\centering
\footnotesize
\setlength{\tabcolsep}{5pt}
\renewcommand{\arraystretch}{1.15}
\caption{Mean $\pm$ std MAE across regions by horizon group, modality configuration, and model. Values are averaged over regions and horizons within each group, after averaging MAE across cross-validation folds. Groups: Short (2, 6, 10 weeks), Medium (14, 18, 22 weeks), Long (26, 30, 34 weeks). Bold: best configuration per model-group pair. Shaded: best across all models. Naive-12Mean and ARIMA are univariate price-only baselines reported as single summary rows. Modality codes: P=prices, C=counts, S=sentiment, B=SAR backscatter, I=interest rates, G=global context.}
\label{tab:grouped_all_models_by_tag_compact}
\begin{tabular}{l l r r r r r r}
\toprule
\textbf{Model} & \textbf{Group} & \textbf{P} & \textbf{PC} & \textbf{PCS} & \textbf{PCSB} & \textbf{PCSBI} & \textbf{PCSBIG} \\
\midrule
Ridge & Short  & \bestoverall{\bestmae{1.988}{1.269}} & \mae{2.248}{1.448} & \mae{2.260}{1.435} & \mae{2.446}{1.544} & \mae{2.608}{1.634} & \mae{2.912}{1.836} \\
      & Medium & \bestmae{3.426}{1.704} & \mae{4.005}{1.968} & \mae{3.987}{1.829} & \mae{4.254}{1.818} & \mae{4.428}{1.818} & \mae{4.687}{1.820} \\
      & Long   & \bestmae{4.373}{2.130} & \mae{5.135}{2.525} & \mae{5.055}{2.208} & \mae{5.097}{2.220} & \mae{5.324}{2.065} & \mae{5.278}{1.980} \\
\addlinespace[2pt]
RF    & Short  & \mae{2.905}{1.771} & \mae{2.887}{1.738} & \mae{2.876}{1.707} & \bestmae{2.796}{1.560} & \mae{2.949}{1.666} & \mae{2.839}{1.574} \\
      & Medium & \mae{3.858}{1.884} & \mae{3.687}{1.829} & \mae{3.560}{1.821} & \mae{3.376}{1.686} & \mae{3.484}{1.773} & \bestmae{3.218}{1.557} \\
      & Long   & \mae{4.310}{2.253} & \mae{3.928}{1.936} & \mae{3.738}{1.850} & \mae{3.486}{1.789} & \mae{3.583}{1.902} & \bestmae{3.260}{1.580} \\
\addlinespace[2pt]
KNN   & Short  & \mae{3.104}{1.760} & \mae{3.309}{1.782} & \mae{3.224}{1.712} & \mae{3.161}{1.611} & \mae{3.090}{1.609} & \bestmae{3.022}{1.566} \\
      & Medium & \mae{4.031}{2.077} & \mae{3.749}{1.916} & \mae{3.463}{1.755} & \mae{3.146}{1.588} & \mae{3.002}{1.618} & \bestmae{2.910}{1.551} \\
      & Long   & \mae{4.476}{2.461} & \mae{4.030}{2.177} & \mae{3.603}{1.876} & \mae{3.185}{1.708} & \mae{3.120}{1.817} & \bestmae{2.929}{1.504} \\
\addlinespace[2pt]
LSTM  & Short  & \bestmae{4.746}{2.956} & \mae{5.258}{3.291} & \mae{6.028}{3.638} & \mae{5.741}{3.429} & \mae{5.615}{3.129} & \mae{5.883}{3.591} \\
      & Medium & \bestmae{5.189}{3.325} & \mae{5.462}{3.404} & \mae{5.893}{3.601} & \mae{5.864}{3.906} & \mae{5.702}{3.615} & \mae{5.603}{3.216} \\
      & Long   & \mae{5.541}{3.594} & \bestmae{5.399}{3.317} & \mae{6.138}{4.030} & \mae{5.958}{4.056} & \mae{5.905}{3.822} & \mae{5.820}{3.853} \\
\addlinespace[2pt]
XGB   & Short  & \mae{3.075}{1.799} & \mae{3.004}{1.780} & \mae{2.935}{1.687} & \mae{2.859}{1.591} & \mae{2.940}{1.648} & \bestmae{2.822}{1.576} \\
      & Medium & \mae{3.955}{1.939} & \mae{3.649}{1.833} & \mae{3.428}{1.775} & \mae{3.248}{1.616} & \mae{3.316}{1.706} & \bestmae{3.056}{1.516} \\
      & Long   & \mae{4.417}{2.330} & \mae{3.976}{2.071} & \mae{3.644}{1.894} & \mae{3.385}{1.780} & \mae{3.493}{1.942} & \bestmae{3.139}{1.530} \\
\addlinespace[2pt]
ARIMA & All & \multicolumn{6}{c}{3.582 (Short) / 4.404 (Medium) / 5.116 (Long)} \\
Naive-12Mean & All & \multicolumn{6}{c}{2.489 (Short) / 3.387 (Medium) / 4.183 (Long)} \\
\bottomrule
\end{tabular}
\end{table*}

We assess whether performance differences represent genuine improvements using nonparametric tests across all 19 regions at long horizons ($h \geq 26$). Table~\ref{tab:significance_tags} reports within-model tests for modality additions. Overall variation across modality configurations is significant for all evaluated learning models, including Ridge ($p=0.00606$), LSTM ($p=8.82\times10^{-4}$), and the nonparametric and ensemble methods ($p<0.001$ for KNN, RF, and XGB). The pattern of gains differs by model. KNN shows the clearest and most consistent multimodal response: adding sentiment yields a significant improvement (\texttt{PC}$\rightarrow$\texttt{PCS}, $p_{\text{corr}}=0.00261$), and adding SAR produces a further strong reduction in error (\texttt{PCS}$\rightarrow$\texttt{PCSB}, $p_{\text{corr}}=5.7\times10^{-5}$). Random Forest benefits significantly from SAR (\texttt{PCS}$\rightarrow$\texttt{PCSB}, $p_{\text{corr}}=0.00134$) and from adding global context (\texttt{PCSBI}$\rightarrow$\texttt{PCSBIG}, $p_{\text{corr}}=0.00483$). XGBoost exhibits significant improvements at multiple stages, including sentiment (\texttt{PC}$\rightarrow$\texttt{PCS}, $p_{\text{corr}}=2.67\times10^{-4}$), SAR (\texttt{PCS}$\rightarrow$\texttt{PCSB}, $p_{\text{corr}}=0.00210$), and global context (\texttt{PCSBI}$\rightarrow$\texttt{PCSBIG}, $p_{\text{corr}}=0.00586$). Ridge improves only when transaction counts are added to prices (\texttt{P}$\rightarrow$\texttt{PC}, $p_{\text{corr}}=0.00483$), with no further significant multimodal gains. LSTM shows a statistically significant improvement only for sentiment addition (\texttt{PC}$\rightarrow$\texttt{PCS}, $p_{\text{corr}}=0.00847$), but remains substantially less accurate than nonparametric methods in Table~\ref{tab:grouped_all_models_by_tag_compact}.

\begin{table}[t]
\centering
\footnotesize
\setlength{\tabcolsep}{5pt}
\renewcommand{\arraystretch}{1.15}
\caption{Within-model significance tests across modality configurations ($h \geq 26$). Friedman tests assess overall variation; Wilcoxon signed-rank tests with Bonferroni correction assess adjacent transitions. ``n.s.'' denotes non-significant after correction.}
\label{tab:significance_tags}
\begin{tabular}{lll}
\toprule
\textbf{Model} & \textbf{Friedman} & \textbf{Significant Transitions (Bonferroni-corrected $p$)} \\
\midrule
Ridge & $p=0.00606$ & \texttt{P}$\rightarrow$\texttt{PC}: 0.00483; others n.s. \\[2pt]
RF    & $p<0.001$ & \texttt{PCS}$\rightarrow$\texttt{PCSB}: 0.00134; \texttt{PCSBI}$\rightarrow$\texttt{PCSBIG}: 0.00483 \\[2pt]
KNN   & $p<0.001$ & \texttt{PC}$\rightarrow$\texttt{PCS}: 0.00261; \texttt{PCS}$\rightarrow$\texttt{PCSB}: $5.7\times10^{-5}$ \\[2pt]
LSTM  & $p=8.82\times10^{-4}$ & \texttt{PC}$\rightarrow$\texttt{PCS}: 0.00847; others n.s. \\[2pt]
XGB   & $p<0.001$ & \texttt{PC}$\rightarrow$\texttt{PCS}: 0.00027; \texttt{PCS}$\rightarrow$\texttt{PCSB}: 0.00210; \texttt{PCSBI}$\rightarrow$\texttt{PCSBIG}: 0.00586 \\
\bottomrule
\end{tabular}
\end{table}

\subsection{Model Diagnostics}

To understand why multimodal inputs improve KNN but not LSTM, we examine model behavior at the longest horizon ($h = 34$). We compare price-only (\texttt{P}) and full multimodal (\texttt{PCSBIG}) configurations using three metrics: percentage MAE reduction, mean fraction of shared nearest neighbors, and baseline MAE. 

\subsubsection{KNN}

Regions with the largest long-horizon improvements under multimodal inputs (e.g., Jumeirah Village Triangle, International City Phase~1, Mudon) exhibit relatively stable nearest-neighbor structures when moving from price-only to full multimodal features. We quantify neighborhood stability by the mean overlap between the $k$ nearest neighbors selected by KNN using price-only features and those selected under the full multimodal representation at $h=34$. Across regions, mean neighbor overlap ranges from approximately 0.49 to 0.73, which reflects heterogeneity in nearest-neighbor selection under multimodal features. Across regions, the association between neighbor overlap and percentage MAE reduction is positive but modest (Pearson $r=0.19$, Spearman $\rho=0.24$). In contrast, baseline price-only error exhibits a somewhat stronger correlation with relative improvement (Pearson $r=0.35$, Spearman $\rho=0.28$), suggesting that multimodal gains depend both on local neighborhood stability and on the underlying predictability of the regional price series. Taken together, these diagnostics indicate that multimodal improvements in KNN result from modifications to nearest-neighbor sets rather than complete replacement, and that neighborhood stability alone is insufficient to explain cross-regional variation in gains.

\subsubsection{LSTM}
To assess whether the LSTM transforms multimodal inputs into predictive representations, we apply linear probes~\cite{alain2018understandingintermediatelayersusing} to frozen hidden states at $h=34$. Each probe is a linear model mapping the final hidden representation to each modality's input features. All modalities exhibit high linear recoverability ($R^2 > 0.94$, ranging from $0.944$ for sentiment to $0.983$ for interest rates), indicating that the LSTM explicitly encodes multimodal inputs rather than discarding or abstracting them. Despite this high-fidelity encoding, the LSTM does not achieve improved long-horizon accuracy. This diagnostic suggests that its underperformance stems from ineffective utilization of encoded signals rather than missing information.

\subsection{Decomposing Signal Contributions}
In this section we expand the additive tag system beyond the six main configurations in Table~\ref{tab:tags_config} to conduct ablations. Tags remain defined as subsets of modalities $\{P,C,S,B,I,G\}$, but instead of evaluating only the single additive chain (e.g., \texttt{PC}$\rightarrow$\texttt{PCS}$\rightarrow$\texttt{PCSB}), we also evaluate additional combinations obtained by adding or removing specific modality blocks. All $\Delta$MAE values in the contribution analysis are averaged across all horizons, regions, and rolling cross-validation folds.

\subsubsection{Marginal Contributions over Transactional Baselines}

\begin{figure*}
\centering
\includegraphics[width=\textwidth]{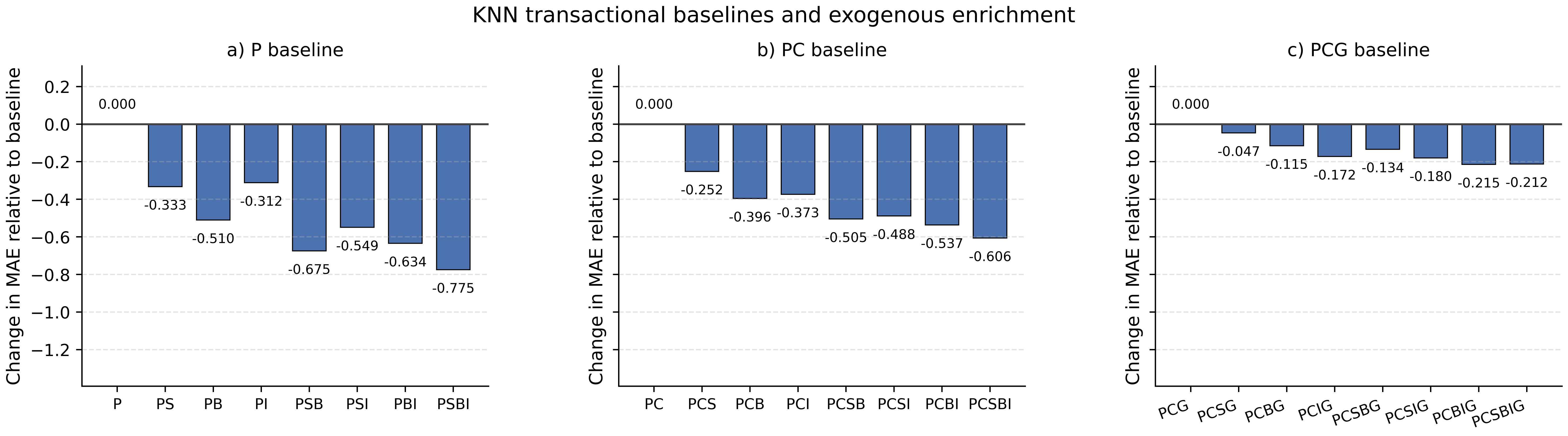}
\caption{Modality contributions relative to transaction baselines (KNN). Panels show incremental additions of sentiment, SAR, and interest rates. Negative $\Delta$MAE indicates improvement; largest gains arise from combined sentiment and SAR.}
\Description{Bar charts showing MAE changes when adding modalities to transaction baselines. All additions show negative delta MAE, with sentiment plus SAR combinations showing largest improvements.}
\label{fig:ablation_core}
\end{figure*}

Figure~\ref{fig:ablation_core} reports changes in MAE when exogenous modalities are added to three transaction-based baselines: prices only (\texttt{P}), prices plus counts (\texttt{PC}), and prices, counts, and global context (\texttt{PCG}). For each baseline, sentiment ($S$), SAR backscatter ($B$), and interest rates ($I$) are added individually and in combination, and $\Delta$MAE is reported relative to the corresponding baseline. Across all baselines, SAR provides the largest single-modality improvement, while sentiment and interest rates yield smaller but consistent gains. The strongest reductions arise when sentiment, SAR, and interest rates are combined. Relative to \texttt{P}, the best configuration reduces MAE by 0.77 (approximately 20\%). Relative to \texttt{PC}, the best enrichment achieves a 0.61 point reduction (16.5\%). When global context is already included (\texttt{PCG}), additional gains are more modest (0.21 points, 6.8\%), indicating diminishing marginal returns once higher-level market structure is incorporated.

\subsubsection{Exogenous-Only Signals}

\begin{figure*}
\centering
\includegraphics[width=\textwidth]{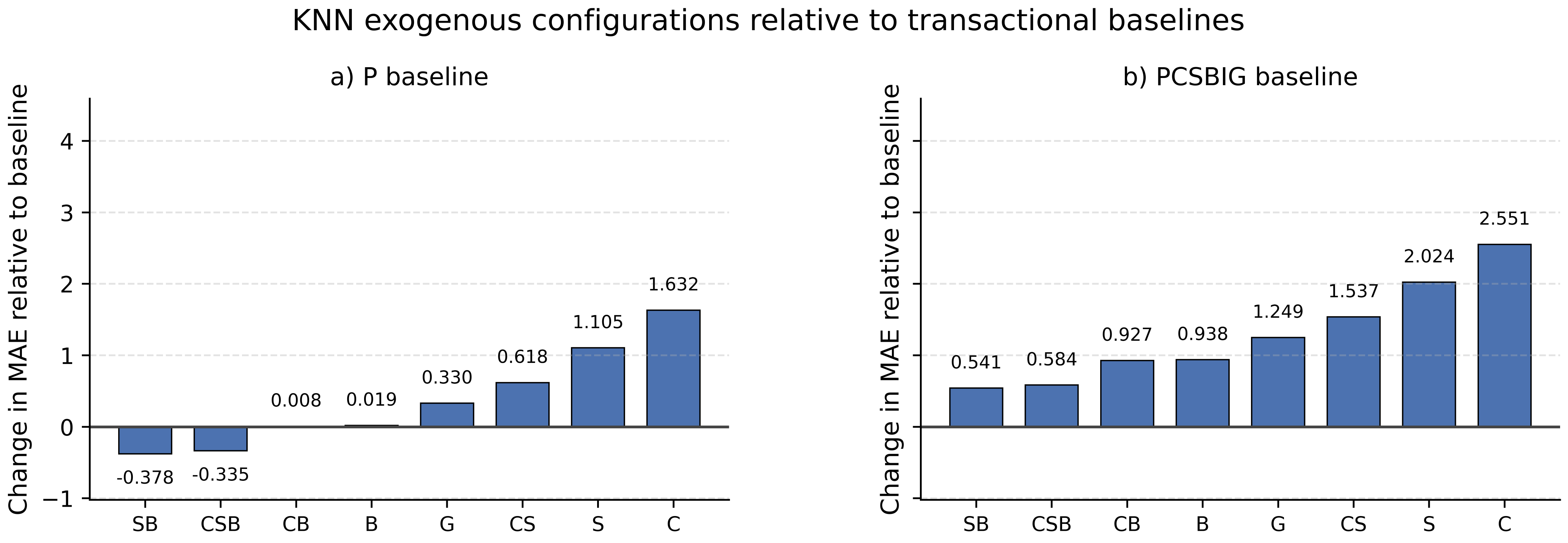}
\caption{Ablations excluding regional price history (KNN). (a) $\Delta$MAE relative to price-only baseline. (b) $\Delta$MAE relative to full multimodal configuration. Sentiment and SAR provide value primarily when combined with transaction data.}
\Description{Two panel figure showing performance of exogenous-only configurations. Single modalities underperform transaction baselines but combinations show modest gains.}
\label{fig:ablation_ext}
\end{figure*}

Figure~\ref{fig:ablation_ext} evaluates configurations that exclude regional price history, using subsets of counts ($C$), sentiment ($S$), SAR ($B$), and global context ($G$). Relative to the price-only baseline, single-modality inputs perform poorly, and neither sentiment nor SAR alone is sufficient to match autoregressive performance. However, combining sentiment and SAR yields moderate improvements over prices, with the \texttt{SB} and \texttt{CSB} configurations reducing MAE by approximately 0.38 and 0.34 points, respectively. Despite these gains, all exogenous-only configurations perform worse than the full multimodal model. Relative to the \texttt{PCSBIG} configuration, errors increase by 0.5 to 2.5 MAE points. This confirms that while exogenous signals can partially substitute for transaction history at long horizons when combined, they cannot replace regional price dynamics, and the strongest performance arises from their integration with transactional data.

\subsection{Alternative Representations of Sentiment and Built Environment}

Beyond establishing that sentiment and remote sensing signals improve long-horizon forecasts, we examine how alternative representations of the same underlying factors affect predictive performance. We focus on two cases: (i) lexical versus semantic representations of news-derived sentiment, and (ii) radar versus optical representations of the built environment.

\subsubsection{Sentiment: Lexical versus Semantic Representations}

We decompose the sentiment signal to assess whether predictive value derives primarily from lexical tone or semantic content. Table~\ref{tab:representation_ablations} (top panel) compares three sentiment configurations for the \texttt{PC}$\rightarrow$\texttt{PCS} transition: NSI-only (lexical), PCA-only (semantic), and the full block that combines both. All comparisons are conducted on long horizons ($h \geq 26$) using paired tests across regions. Lexical NSI features yield a modest but statistically significant reduction in MAE. Aggregate tone alone therefore contains predictive information at long horizons. Semantic PCA features, however, produce a larger and more consistent improvement. Topical structure captured by embeddings provides richer market-relevant signals than polarity alone. The full sentiment block achieves the largest MAE reduction, though its advantage over the PCA-only configuration is incremental. Most of the predictive value therefore arises from semantic components. These results demonstrate that lexical tone contributes marginally to forecast accuracy, but long-horizon gains depend primarily on semantic representations.

\subsubsection{Urban Environment: SAR versus Optical Indices}

\begin{figure}
\centering
\includegraphics[width=0.7\linewidth]{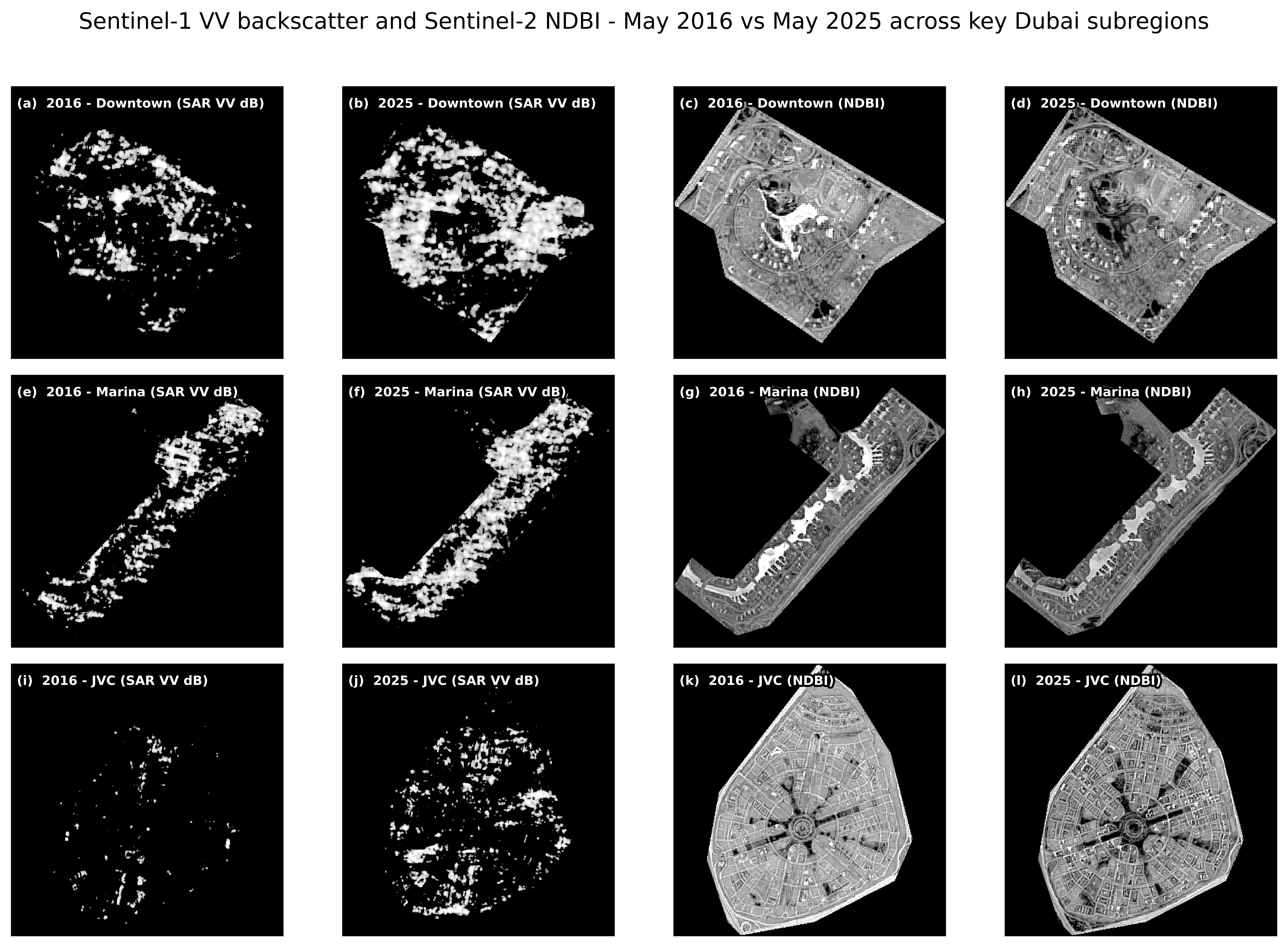}
\caption{SAR and NDBI comparison across three Dubai subregions (2016 vs.\ 2025). SAR backscatter (left columns) shows consistent sensitivity to built-up development, while optical NDBI (right columns) exhibits greater variability.}
\Description{Grid of 12 images showing SAR and NDBI for three regions at two time points, with annotations highlighting development changes.}
\label{fig:sar_ndbi_regions}
\end{figure}

We next compare Sentinel-1 SAR backscatter~\cite{torres_gmes_2012} against Sentinel-2 NDBI~\cite{zha_use_2003,drusch_sentinel-2_2012} as alternative representations of urban-environment dynamics. Figure~\ref{fig:sar_ndbi_regions} provides a qualitative comparison across three major subregions over time. SAR backscatter shows consistent sensitivity to structural changes associated with urban development. NDBI, by contrast, exhibits greater variability due to atmospheric conditions and illumination effects. Table~\ref{tab:representation_ablations} (bottom panel) reports paired one-sided tests that compare SAR and NDBI against no-$B$ baselines at long horizons ($h \geq 26$). For the \texttt{PCS}$\rightarrow$\texttt{PCSB} transition, SAR yields a large and highly significant reduction in MAE (mean $\Delta$MAE $=-0.43$, $p=1.8\times10^{-4}$). NDBI produces no improvement under this configuration.

\begin{table}[t]
\centering
\footnotesize
\setlength{\tabcolsep}{10pt}
\renewcommand{\arraystretch}{1.15}
\caption{Representation ablations for sentiment and built-environment modalities (KNN, $h \geq 26$). Reported values are mean $\Delta$MAE relative to the no-modality baseline for each transition, with one-sided paired tests.}
\label{tab:representation_ablations}
\begin{tabular}{lcc}
\toprule
\textbf{Feature Variant} & \textbf{Mean $\Delta$MAE} & \textbf{$p$-value} \\
\midrule
\multicolumn{3}{l}{\textit{Sentiment (\texttt{PC}$\rightarrow$\texttt{PCS})}} \\[2pt]
\quad NSI only (lexical)    & $-0.124$ & $0.0348$ \\
\quad PCA only (semantic)   & $-0.193$ & $0.0069$ \\
\quad Full sentiment block  & $-0.264$ & $0.0047$ \\[6pt]
\multicolumn{3}{l}{\textit{Built environment (\texttt{PCS}$\rightarrow$\texttt{PCSB})}} \\[2pt]
\quad SAR backscatter       & $-0.429$ & $1.8\times 10^{-4}$ \\
\quad Optical NDBI          & $+0.058$ & $0.94$ (n.s.) \\
\bottomrule
\end{tabular}
\end{table}

\subsection{Out-of-Sample Forecast Behavior}

\begin{figure*}
\centering
\includegraphics[width=0.7\textwidth]{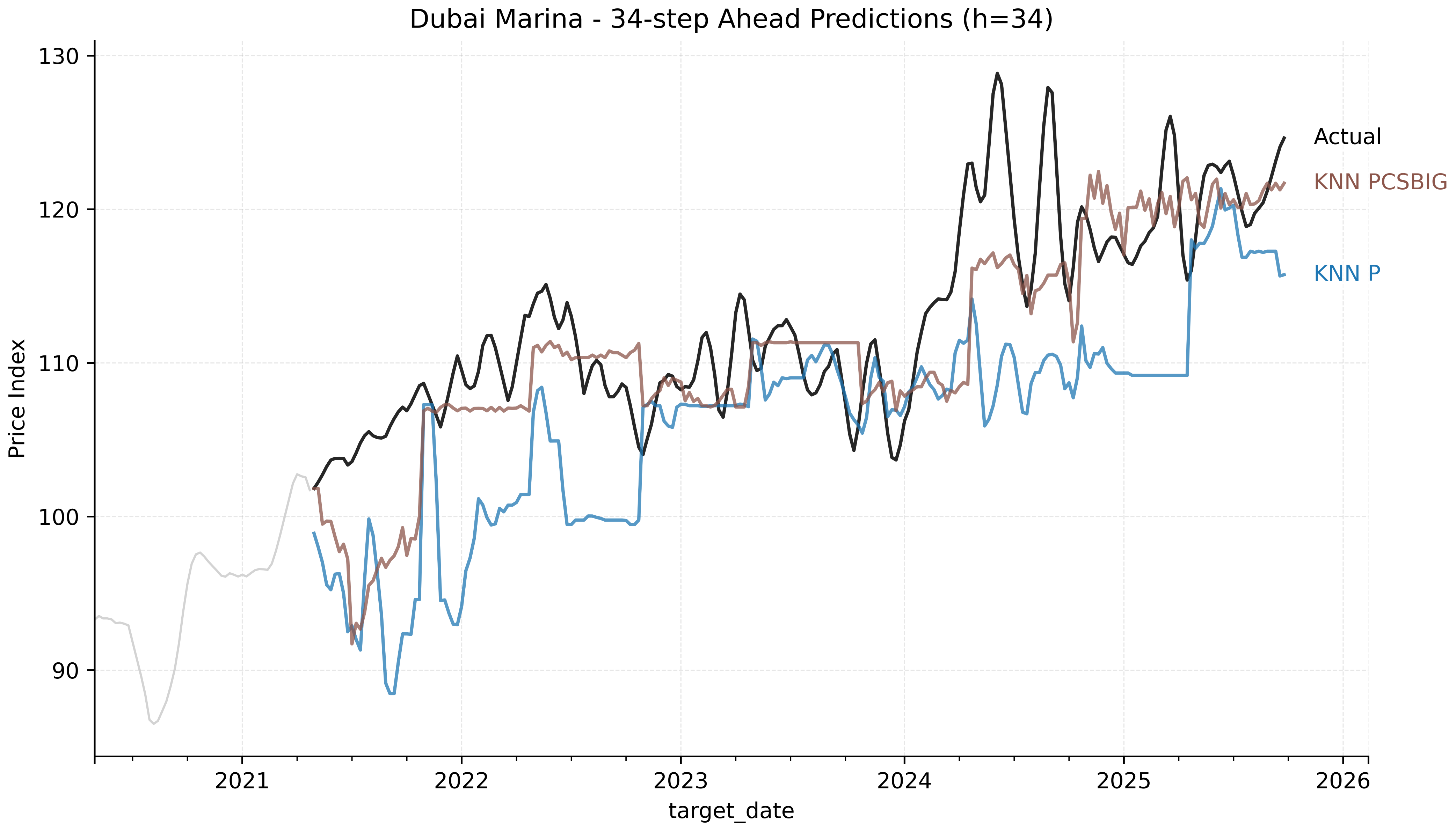}
\caption{Out-of-sample 34-week forecasts for the Dubai Marina price index (2022--2025). The price-only model displays higher volatility and delayed trend response, while the multimodal configuration produces smoother forecasts that track realized values more closely.}
\Description{Time series plot comparing price-only and multimodal forecasts against realized Dubai Marina price index values. Multimodal forecasts exhibit reduced volatility and improved medium-term trend alignment.}
\label{fig:dubai_marina_predictions}
\end{figure*}

Figure~\ref{fig:dubai_marina_predictions} presents out-of-sample predictions for the Dubai Marina sub-city price index at 34 weeks ahead. The forecast series is constructed by stitching non-overlapping validation windows. The price-only model exhibits pronounced oscillations and delayed responses around turning points. The multimodal configuration, in contrast, tracks the realized series more smoothly over the 2022--2025 period. It captures medium-term acceleration and deceleration phases more consistently than the price-only model. Improvements are particularly visible during high-volatility intervals. Multimodal forecasts reduce persistent gaps relative to observed prices in these periods. These results are consistent with the autoregressive limitations and multimodal stabilization effects found earlier.

\section{Conclusion}\label{sec:conclusion}

This study asked whether satellite imagery and news sentiment can reveal real estate market dynamics that transaction records alone cannot capture. Using weekly price indices for Dubai sub-city regions constructed from over 350,000 transactions, we find that the answer is affirmative but horizon dependent. At short horizons (up to 10 weeks ahead), transaction history alone performs strongly as autoregressive patterns dominate. Beyond 14 weeks, however, price-driven patterns decay and non-traditional signals provide significant gains that preserve predictability. At long horizons (26--34 weeks ahead), integrating Sentinel-1 SAR backscatter and news sentiment reduces forecast error by roughly one third compared to price-only baselines, with improvements statistically significant across regions.

Three findings are particularly notable. First, nonparametric learners (KNN and Random Forest) consistently outperform both linear models and LSTM in this weekly sample regime. This suggests that the forecasting challenge depends more on feature engineering than architectural complexity. Second, semantic news embeddings carry more predictive value than lexical sentiment scores. Topical structure matters more than aggregate positive-negative tone. Third, SAR backscatter significantly outperforms optical NDBI as a built-environment proxy, likely because radar captures surface changes from construction activity that precede transaction records and does this more reliably than optical imagery over time.

As a result, we show that sub-city dynamics are predictable at strategically relevant horizons and frequency when transaction records are combined with signals reflecting physical development and market sentiment. The modular framework and systematic ablations we introduced establish reproducible benchmarks for multimodal index forecasting and clarify when non-traditional signals add value. The framework's applicability beyond Dubai remains to be tested. Markets with sparse reporting or slower development may exhibit different signal dynamics. For future work, cross-regional dependencies modeled through graph representations offer a natural extension \cite{zhang_mugrep_2021, uygun_financial_2025}. A second direction to explore is topic-aware sentiment models tailored to real estate discourse as a path toward more interpretable market factors.

\section*{Data and Code Availability}
The code repository and all processed features used in modeling (the weekly aligned feature matrices) will be released upon acceptance. All raw data sources used in this study are publicly accessible: transaction records from the Dubai Land Department (Open Data License), news metadata from the GDELT Project (article text retrieved independently is not redistributed, instead aggregated sentiment features are provided), Sentinel-1 and Sentinel-2 satellite imagery from the Copernicus Programme, and interest rates from the Central Bank of the UAE.

\bibliographystyle{plainnat}
\bibliography{RealEstateDubai}

\end{document}